\documentclass[11pt,a4paper]{article}
\usepackage{acl2015}
\usepackage{times}
\usepackage{url}
\usepackage{latexsym}
\usepackage{algorithm} 
\usepackage{algorithmic}
\usepackage{bm}
\usepackage{graphicx}
\usepackage{enumitem}
\usepackage{multirow}
\usepackage{todonotes}
\usepackage{tikz}
\usetikzlibrary{arrows}
\usepackage{caption}
\usepackage{subcaption}
\newcommand{\rulesep}{\unskip\ \vrule\ }

 \title{Parser for Abstract Meaning Representation using Learning to Search}
\author{ 
  \textbf{Sudha Rao}${}^{1,3}\thanks{The first two authors contributed equally to this work.}$\textnormal{,} \textbf{Yogarshi Vyas}${}^{1,3*}$\textnormal{,} \textbf{Hal Daum\'e III}${}^{1,3}$\textnormal{,} \textbf{Philip Resnik}${}^{2,3}$\\
  ${}^1$Computer Science,
  ${}^2$Linguistics,
  ${}^3$UMIACS\\
  University of Maryland\\
  raosudha@cs.umd.edu, yogarshi@cs.umd.edu, hal@cs.umd.edu, resnik@umd.edu
  }
\date{}

\begin{document}
\maketitle
\begin{abstract}
We develop a novel technique to parse English sentences into Abstract Meaning Representation (AMR) using \textsc{searn}, a Learning to Search approach, by modeling the concept and the relation learning in a unified framework. We evaluate our parser on multiple datasets from varied domains and show an absolute improvement of 2\% to 6\% over the state-of-the-art. Additionally we show that using the most frequent concept gives us a baseline that is stronger than the state-of-the-art for concept prediction. We plan to release our parser for public use.

\end{abstract}
\section{Introduction}
\begin{figure}[h!]
\begin{tikzpicture}[->,>=stealth',shorten >=1pt,auto,node distance=2.5cm,
  thick,main node/.style={font=\footnotesize,circle,draw}]
 
  \node[main node] (1){read-01};
  \node[main node] (2) [below left of=1] {i};
  \node[main node] (3) [below right of=1] {book};
  \node[main node] (4) [below left of=3] {name};
  \node[main node] (5) [below right of=3] {forest};
  \node[main node] (6) [below left of=4] {Stories};
  \node[main node] (7) [below of=4] {from};
  \node[main node] (8) [below right of=4] {Nature};
  \path[every node/.style={font=\sffamily\small}]
    (1) edge node [bend left, left] {ARG0} (2)
    (1) edge node [bend right] {ARG1} (3)
    (3) edge node [bend left] {name} (4)
    (3) edge node [bend right] {topic} (5)
    (4) edge node [bend left, left] {op1} (6)
    (4) edge node [below, left] {op2} (7)
    (4) edge node [bend right] {op3} (8)
        ;
\end{tikzpicture}
\caption{AMR graph for the sentence \textit{``I read a book, called Stories from Nature, about the forest.''}}
\label{fig:eg_amr}
\end{figure}
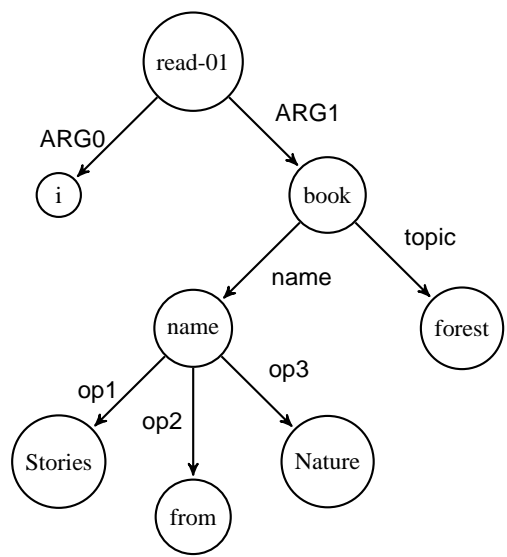

Abstract Meaning Representation \cite{banarescu2013abstract} is a semantic representation which is a rooted, directed, acyclic graph where the nodes represent concepts (words, PropBank \cite{DBLP:journals/coling/PalmerKG05} framesets or special keywords) and the edges represent relations between these concepts. Figure \ref{fig:eg_amr} shows the complete AMR for a sample sentence.

The key motivation behind developing AMR was to have a comprehensive and broad-coverage semantic formalism that puts together the best insights from a variety of semantic annotations (like named entities, co-reference, semantic relations, discourse connectives, temporal entities, etc.) in a way that would enable it to have the same kind of impact that syntactic treebanks have on natural language processing tasks. Currently, there are approximately 20,000 sentences which have been annotated with their AMRs, but for such a representation to be useful for almost any NLP task, a larger set of annotations would be needed. Algorithms that can perform automatic semantic parsing of sentences into AMR can help alleviate the problem of paucity of manual annotations. 

Automatic semantic parsing for AMR is still in a nascent stage. There have been two published approaches for automatically parsing English sentences into AMR. Flanigan et al. \shortcite{DBLP:conf/acl/FlaniganTCDS14} use a semi-Markov model to first identify the concepts, and then find a maximum spanning connected subgraph that defines the relations between these concepts. The other approach \cite{wangtransition} uses a transition-based algorithm to convert the dependency representation of a sentence to its AMR. 

\begin{figure*}[!t]
\fbox{
\begin{subfigure}[b]{1\textwidth}
\centering
\begin{tikzpicture}[->,>=stealth',shorten >=1pt,auto,node distance=1.75cm,
  thick,main node/.style={font=\large}]
  \node[main node] (1) {I};
  \node[main node] (2) [right of=1] {read};
   \node[main node] (3) [right of=2] {a};
  \node[main node] (4) [right of=3] {book};
  \node[main node] (5) [right of=4] {called};
  \node[main node] (6) [right of=5] {Stories};
  \node[main node] (7) [right of=6] {from};
    \node[main node] (8) [right of=7] {Nature};  
\end{tikzpicture}
 \begin{tikzpicture}[->,>=stealth',shorten >=1pt,auto,node distance=1.75cm,
  thick,main node/.style={font=\footnotesize,circle,fill=pink!20,draw}, null node/.style={font=\footnotesize,circle,draw}]
  \node[main node] (1) {i};
  \node[main node] (2) [right of=1] {read-01};
   \node[main node] (3) [right of=2] {NULL};
  \node[main node] (4) [right of=3] {book};
  \node[null node] (5) [right of=4] {NULL};
  \node[null node] (9) [above of=5] {call-01};
  \node[null node] (10) [below of=5] {called};
  \node[null node] (6) [right of=5] {Stories};
   \node[null node] (11) [above of=6] {story};
   \node[null node] (12) [below of=6] {NULL};
  \node[null node] (7) [right of=6] {from};
   \node[null node] (13) [above of=7] {NULL};
   \node[null node] (8) [right of=7] {Nature};
   \node[null node] (14) [above of=8] {NULL};
\end{tikzpicture}
\caption{Concept prediction stage: Shaded nodes indicate predicted concepts (Current state). The middle row represents the oracle action. Other rows represents other possible actions.}
\label{fig:conceptpred}
\end{subfigure}
}
\fbox{
\begin{subfigure}[b]{0.25\textwidth}
\centering
\begin{tikzpicture}[->,>=stealth',shorten >=1pt,auto,node distance=1.5cm,
  thick, main node/.style={font=\footnotesize,circle,draw}, rel node/.style={font=\footnotesize,circle,fill=pink!20,draw}, root node/.style={font=\footnotesize,circle,fill=red!20,draw}]
  
  \node[rel node] (2) {r};
  \node[rel node] (1) [ below left of=2]{i};
  \node[rel node] (3) [ below right of=2]{b};
        \path[every node/.style={font=\small}]
    (2) edge node [bend left, above left] {ARG0} (1)
    ;
\end{tikzpicture}
\caption{Sample current state for relation prediction}
\end{subfigure}
\rulesep
\begin{subfigure}[b]{0.75\textwidth}
\centering
\begin{tikzpicture}[->,>=stealth',shorten >=1pt,auto,node distance=1.5cm,
  thick, main node/.style={font=\footnotesize,circle,draw}, rel node/.style={font=\footnotesize,circle,draw}, root node/.style={font=\footnotesize,circle,draw}]
  
  \node[rel node] (5) [ right of=3]{i};
  \node[root node] (4) [above right of=5] {r};
  \node[rel node] (6) [ below right of=4]{b};
  \node[rel node] (7) [ right of=6]{i};
  \node[root node] (8) [above right of=7] {r};
  \node[rel node] (9) [ below right of=8]{b};
    \node[rel node] (10) [ right of=9]{i};
  \node[root node] (11) [above right of=10] {r};
  \node[rel node] (12) [ below right of=11]{b};
    \path[every node/.style={font=\small}]
    (4) edge node [bend left, above left] {ARG0} (5)
    (6) edge node [bend left] {ARG1} (5)
    (8) edge node [bend left, above left] {ARG0} (7)
    (9) edge node [bend left] {mod} (7)
    (11) edge node [bend left, above left] {ARG0} (10)
    
    ;
\end{tikzpicture}
\caption{Three possible actions given the current state for relation prediction, the last one being the true relation i.e. no edge}
\label{fig:relpred}
\end{subfigure}
}
\caption{Using \textsc{SEARN} for AMR parsing}
\label{fig:runeg}
\end{figure*}
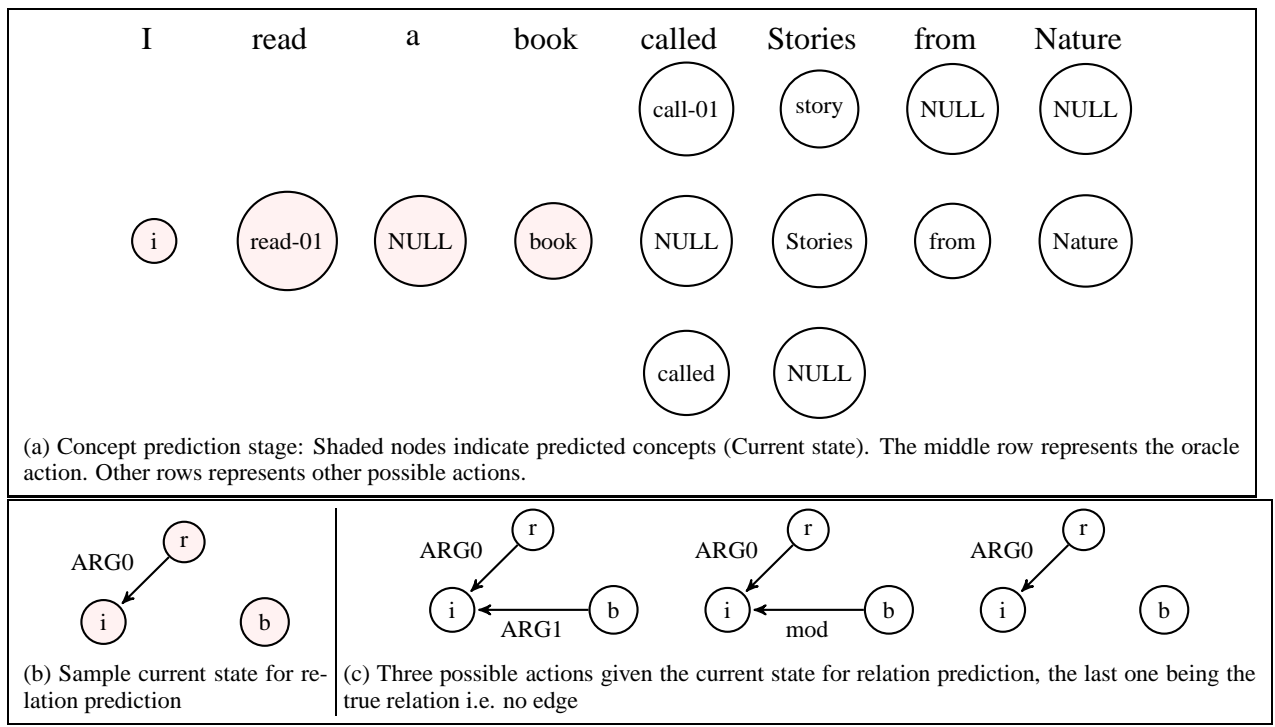

In this work, we develop a novel technique for AMR parsing that uses \textsc{searn} \cite{daume2009search}, a Learning to Search (L2S) algorithm. \textsc{searn} and other L2S algorithms have proven to be highly effective for tasks like part-of-speech tagging, named entity recognition \cite{daume2014efficient}, and for even more complex structured prediction tasks like coreference resolution \cite{ma2014prune} and dependency parsing \cite{DBLP:conf/emnlp/HeDE13}. Using \textsc{searn} allows us to model the learning of concepts and relations in a unified framework which aims to minimize the loss over the entire predicted structure, as opposed to minimizing the loss over concepts and relations in two separate stages, as is done by Flanigan et al \shortcite{DBLP:conf/acl/FlaniganTCDS14}.

There are three main contributions of this work. Firstly, we provide a novel algorithm based on \textsc{searn} to parse sentences into AMRs. Additionally, our parser extracts possible `candidates' for the right concepts and relations from the entire training data, but only uses smaller sentences to train the learning algorithm. This is important since AMR annotations are easier to obtain for smaller sentences. Secondly, we evaluate our parser on datasets from various domains, unlike previous works, which have been restricted to newswire text. We observe that our parser performs better than the existing state-of-the-art parser, with an absolute improvement of 2 to 6 \% over the different datasets. Finally, we show that using the most frequently aligned concept for each word in the sentence (as seen in the training data) as the predicted concept, proves to be a strong baseline for concept prediction. This baseline does better than existing approaches, and we show that our parser performs as well as the baseline at this part of the task in some datasets, and even better in some others.

The rest of this paper is organized as follows. In the next section, we briefly review \textsc{searn} and explain its various components with respect to our AMR parsing task. Section \ref{sec:methods} describes our main algorithm along with the strategies we use to deal with the large search space of the search problem. We then describe our experiments and results (Section \ref{sec:exp}).

\section{Using \textsc{searn}} 
\label{sec:searn}

The task of generating an AMR given a sentence is a structured prediction task where the structure that we are trying to predict is a singly rooted, connected directed graph with concepts (nodes) and relations (edges). In this work, we design an AMR parser that learns to predict this structure using \textsc{searn}. \textsc{searn} solves complex structured prediction problems by decomposing it into classification problems. It does so by decomposing the structured output, $y$, into a sequence of decisions $y_1, y_2, ..., y_m$ and then using a classifier to make predictions for each component in turn, allowing for dependent predictions. We decompose the AMR prediction problem into the three problems of predicting the concepts of the AMR, predicting the root and then predicting the relations between the predicted concepts (explained in more detail under section \ref{sec:methods}). Below, we explain how we use \textsc{searn}, with reference to a running example in Figure \ref{fig:runeg}.

\textsc{searn} works on the notion of a policy which can be defined as ``what is the best next action ($y_i$) to take''  in a search space given the current state ($s = (x, y_1, y_2, .., y_{i-1})$), where $x$ is the input. For our problem, a state during the concept prediction phase is defined as the concepts predicted for a part of the input sentence. Similarly, a state during the relation prediction phase is defined as the set of relations predicted for certain pairs of concepts obtained during the concept prediction stage. (In Figure \ref{fig:conceptpred} (concept prediction), the current state corresponds to the concepts \{`i', `read-01', `book'\} predicted for a part of the sentence. In Figure \ref{fig:relpred} (relation prediction), the current state corresponds to the relation `ARG0' predicted between `r' and `i' )

At training time, \textsc{searn} operates in an iterative fashion. It starts with some initial policy and given an input $x$, makes a prediction $y$ = $y_1, y_2, ..., y_m$ using the current policy. For each prediction $y_i$ it generates a set of cost-sensitive multi-class classification examples each of which correspond to a possible action ($a$) the algorithm can take given the current state. Each example can be defined using local features and features that depend on previous predictions. The possible set of next actions in our concept prediction phase corresponds to the set of possible concepts the next word can take. The possible set of next actions in our relation prediction phase corresponds to the set of possible relations the next pair of concepts can take. (In Figure \ref{fig:conceptpred} (concept prediction), the next action is assigning one of  \{`call-01', `called', \textsc{null}\} to the word `called'. In Figure \ref{fig:relpred} (relation prediction), the next action is assigning one of  \{`ARG1', `mod', \textsc{no-edge}\} to the pair of concept `b' and `i'). 

During training, \textsc{searn} has access to an ``oracle'' policy which gives the true best action ($a^*$) given the current state . Our oracle returns the correct concept and relation labels in the concept prediction and relation prediction phase respectively. (In Figure \ref{fig:conceptpred} (concept prediction), the oracle will return \textsc{null} and in Figure \ref{fig:relpred} (relation prediction), the oracle will return \textsc{no-edge}). \textsc{searn} then calculates the loss between $a$ and $a^*$ using a pre-specified loss function. It then computes a new policy based on this loss and interpolates it with the current policy to get an updated policy, before moving on to the next iteration. 

At test time, predictions are made greedily using the policy learned during training time.

\section{Methodology}
\label{sec:methods}

\begin{table*}[tp]
\begin{center}
\begin{tabular}{|c|p{10cm}|}
\hline
\textbf{Feature label}	&	\textbf{Description }	\\
\hline
$w_{i-2}, w_{i-i}, w_{i}, w_{i+1}, w_{i+2}$	&	Words in $s_i$ and context	\\\hline
$p_{i-2}, p_{i-i}, p_{i}, p_{i+1}, p_{i+2}$	&	POS tags of words in $s_i$ and context \\\hline
$NE_i$							&	Named entity tags for words in $s_i$ \\\hline	
$s_i$							&	Binary feature indicating whether $w_i$ is(are) stopword(s) \\\hline
$dep_i$							&  	All dependency edges originating from words in $w_i$ \\\hline

$b_{c}$							&	Binary feature indicating whether $c$ is the most frequently aligned concept with $s_i$ or not \\\hline
$c_{i-2}, c_{i-1}$					&	Predicted concepts for two previous spans \\\hline
$c$								&	Concept label and its conjunction with all previous features \\ \hline
$frame_i$ and $sense_i$				& 	If the label is a PropBank frame (e.g. `see-01', use the frame (`see') and the sense(`01') as additional features.\\\hline
\end{tabular}
\caption{Concept prediction features for span $s_i$ and concept label $c_i$}
\label{tab:featsconcept}
\end{center}
\end{table*}

\begin{table*}[!t]
\begin{center}
\begin{tabular}{|c|p{10cm}|}
\hline
\textbf{Feature label}	&	\textbf{Description }	\\
\hline
$c_i, c_j, c_i \land c_j$	&	The two concepts and their conjunction	\\\hline
$w_i, w_j, w_i \land w_j$	&	Words in the corresponding spans and their conjunction \\\hline
$p_i, p_j, p_i \land p_j$	&	POS tags of words in spans and their conjunction \\\hline	
$dep_{ij}$					&	All dependency edges with tail in $w_i$ and head in $w_j$ \\\hline
$dir$					&  	Binary feature which is true iff $i < j$ \\\hline
$r$						& 	Relation label and its conjunction with all other features \\\hline
\end{tabular}
\caption{Relation prediction features for concepts $c_i$ and $c_j$ and relation label $r$}
\label{tab:featsrelation}
\end{center}
\end{table*}

\begin{table*}[!t]
\begin{center}
\begin{tabular}{|c|p{10cm}|}
\hline
\textbf{Feature label}	&	\textbf{Description }	\\
\hline

$c_i$				&	Concept label. If the label is a PropBank frame (e.g. `see-01', use the frame (`see') and the sense(`01') as additional features.\\\hline
$w_{i}$				&	Words in $s_i$, i.e. the span corresponding to $c_i$ 	\\\hline
$p_i$				&	POS tags of words in $s_i$ \\\hline
$is\_dep\_root_i$		&	Binary feature indicating whether one of the words in $s_i$ is the root in the dependency tree of the sentence \\\hline
\end{tabular}
\caption{Root prediction features for concept $c_i$}
\label{tab:featsroot}
\end{center}
\end{table*}

\subsection{Learning technique}
\label{subsec:learning}

 \begin{algorithm}[!h]
\caption{}
\label{algo:learning}
\begin{algorithmic}[1]

\FOR{\texttt{$each \: span \: s_i$}} 
\STATE \texttt{$c_i = predict\_concept(s_i)$}
\ENDFOR
\STATE \texttt{$c_{root} = predict\_root([c_1, ..., c_n])$}
\FOR{\texttt{$each \: concept \: c_i$}} 
\FOR{\texttt{$each \: j < i$}}
 \STATE \texttt{$r_{(i,j)} = predict\_relation(c_i, c_j)$}
  \STATE \texttt{$r_{(j,i)} = predict\_relation(c_j, c_i)$}
\ENDFOR
\ENDFOR

\end{algorithmic}
\end{algorithm}

We use \textsc{searn} as described in section \ref{sec:searn} to learn a model that can successfully predict the AMR $y$ for a sentence $x$. The sentence $x$ is composed of a sequence of spans $(s_1, s_2, ..., s_n)$ each of which can be a single word or a span of words (We describe how we go from a raw sentence to a sequence of spans in Section \ref{subsec:preproc}). Given that our input has $n$ spans, we first decompose the structure into a sequence of $n^2+1$ predictions $\textbf{D} = (\textbf{C}, \textbf{ROOT}, \textbf{R})$, where

$\textbf{C} = c_1, c_2,  ..., c_n$ - where $c_i$ is the concept predicted for span $s_i$

\textbf{ROOT} is the decision of choosing one of the predicted concepts as the root ($c_{root}$) of the AMR

$\textbf{R} = r_{2,*}, r_{*,2}, r_{3,*}, r_{*,3}, ..., r_{n,*}, r_{*,n} $ - where $r_{i,*}$ are the predictions for the directed relations from  $c_i$ to $c_j$ $\forall j < i$, and $r_{*,i}$ are the predictions for the directed relations from  $c_j$ to $c_i$ $\forall j < i$. We constrain our algorithm to not predict any incoming relations to $c_{root}$.

During training time, the possible set of actions for each prediction is given by the $k$-best list, which we will describe in Section \ref{subsec:kbest}. We use Hamming Loss as our loss function. Under Hamming Loss, the oracle policy is simply choosing the right action for each prediction. Since this loss is defined on the entire predicted output, the model learns to minimize the loss for concepts and relations jointly.

Algorithm \ref{algo:learning} describes the sequence of predictions to be made in our problem. We learn three different policies corresponding to each of the functions $predict\_concept$, $predict\_root$ and $predict\_relation$. The learner in each stage uses features that depend on predictions made in the previous stages. Tables \ref{tab:featsconcept}, \ref{tab:featsrelation} and \ref{tab:featsroot} describe the set of features we use for the concept prediction, relation prediction and root prediction stages respectively.

\subsection{Selecting $k$-best lists}
\label{subsec:kbest}
For predicting the concepts and relations using \textsc{Searn}, we need a candidate-list (possible set of actions) to make predictions from. 

\textbf{Concept candidates:} For a span $s_i$, the candidate-list of concepts, CL-CON$_{s_i}$ is the set of all concepts that were aligned to $s_i$ in the training data. If $s_i$ has not been seen in the training data, CL-CON$_{s_i}$ consists of the lemmatized span, PropBank frames (for verbs) obtained using the Unified Verb Index \cite{schuler2005verbnet} and the \textsc{null} concept.

\textbf{Relation candidates:} The candidate list of relations for a relation from concept $c_i$ to concept $c_j$, CL-REL$_{ij}$, is the union of the following three sets:

\begin{itemize}[noitemsep]

\item$pairwise_{i,j}$ - All directed relations from $c_i$ to $c_j$ when $c_i$ and $c_j$ occurred in the same AMR,

\item$outgoing_i$ - All outgoing relations from $c_i$, and

\item $incoming_j$ - All incoming relations into $c_j$.

\end{itemize}

In the case when both $c_i$ and $c_j$ have not been seen in the training data, CL-REL$_{ij}$ consists of all relations seen in the training data. In both cases, we also provide an option \textsc{no-edge} which indicates that there is no relation between $c_i$ and $c_j$.

\subsection{Pruning the search space}
\label{subsec:depfiltering}
To prune the search space of our learning task, and to improve the quality of predictions, we use two observations about the nature of the edges of the AMR of a sentence, and its dependency tree, within our algorithm.

First, we observe that a large fraction of the edges in the AMR for a sentence are between concepts whose underlying spans (more specifically, the words in these underlying spans) are within two edges of each other in the dependency tree of the sentence. Thus, we refrain from calling the $predict\_relation$ function in Algorithm \ref{algo:learning} between concepts $c_i$ and $c_j$ if each word in $w_i$ is three or more edges away from all words in $w_j$ in the dependency tree of the sentence under consideration, and vice versa. This implies that there will be no relation $r_{ij}$ in the predicted AMR of that sentence. This doesn't affect the number of calls to $predict\_relation$ in the worst case ($n^2-n$, for a sentence with $n$ spans), but practically, the number of calls are far fewer. Also, to make sure that this method does not filter out too many AMR edges, we calculated the percentage of AMR edges that are more than two edges away in dependency tree. We found this number to be only about 5\% across all our datasets. 

Secondly, and conversely, we observe that for a large fraction of words which have a dependency edge between them, there is an edge in the AMR between the concepts corresponding to those two words. Thus, when we observe two concepts $c_i$ and $c_j$ which satisfy this property, we force our $predict\_relation$ function to assign a relation $r_{ij}$ that is not \textsc{Null}.

\subsection{Training on smaller sentences}
\label{subsec:smallersent}

For a sentence containing $n$ spans, Algorithm \ref{algo:learning} has to make $n^2$ predictions in the worst case, and this can be inhibitive for	 large values of $n$. To deal with this, we use a parameter to indicate a cut-off on the length of a sentence ($C$), and only use sentences whose length (number of spans) is less than or equal to $C$. This parameter can be varied based on the size of the training data and the distribution of the length of the sentences in the training data. Setting a higher values of $C$ will cause the model to use more sentences for training, but spend longer time, whereas lower values will train quickly on fewer sentences. In our experiments, a $C$-value between $10$ and $15$ gave us the best balance between training time, and number of examples considered.

\section{Experiments and Results}
\label{sec:exp}

\subsection{Dataset and  Method of Evaluation}
\label{subsec:data}
We use the publicly available AMR Annotation Release 1.0 (LDC2014T12) corpus for our experiments. This corpus consists of datasets from varied domains such as online discussion forums, blogs, and newswire, with about 13,000 sentence-AMR pairs. Previous works have only used one of these datasets for evaluation (proxy), but we evaluate our parser on all of them. Additionally, we also use the freely available AMRs for \textit{The Little Prince}, (lp) \footnote{\url{http://amr.isi.edu/download.html}} which is from a more literary domain. All datasets have a pre-specified training and test split (Table \ref{tab:data}).

As stated earlier (Sections \ref{subsec:kbest} and \ref{subsec:smallersent}), we use the entire training set to extract the candidate lists for concept prediction and relation prediction, but train our learning algorithm on only a subset of the sentence-AMR pairs in the training data, which is obtained by selecting sentences having less than a fixed number of spans ($C$, set to 10 for all our experiments). Table \ref{tab:data} also mentions the number of sentences in each training dataset that are of length $\leq C$ (column \textbf{Training} $(\boldsymbol{\leq C}$)).
\begin{table}[!h]
\begin{center}
\begin{tabular}{|c|c|c|c|}

\hline
\textbf{Dataset}	&	\textbf{Training}		&	\textbf{Training} $(\boldsymbol{\leq C}$)	&	\textbf{Test}	\\\hline

bolt			&		1061			&	119								&	133			\\\hline
proxy		&		6603			&	1723								&	823			\\\hline
xinhua		&		741			&	115								&	86			\\\hline
dfa			&		1703			&	438								&	229			\\\hline
lp			&		1274			&	584								&	173			\\\hline

\end{tabular}
\caption{Dataset statistics. All figures represent number of sentences.}
\label{tab:data}
\end{center}
\end{table}

We compare our results against those of the JAMR parser \footnote{\url{https://github.com/jflanigan/jamr}} of Flanigan et. al \shortcite{DBLP:conf/acl/FlaniganTCDS14} \footnote{The transition-based parser by Wang et al. \cite{} is newer, but the latest release of JAMR performs better, hence we do not compare against the former.}. We run the parser with the configuration that is specified to give the best results. 

The evaluation of predicted AMRs is done using Smatch \cite{smatch} \footnote{\url{http://amr.isi.edu/download/smatch-v2.0.tar.gz}}, which compares two AMRs using precision, recall and $F_1$. Additionally, we also evaluate how good we are at predicting the concepts of the AMRs, by calculating precision, recall and $F_1$ against the gold-concepts that are aligned to the induced spans during test time.

\subsection{Preprocessing}
\label{subsec:preproc}

\textbf{JAMR Aligner}: The training data for AMR parsing consists of sentences paired with corresponding AMRs. To convert a raw sentence into a sequence of spans (as required by our algorithm), we obtain alignments between words in the sentence and concepts in the AMR using the automatic aligner of JAMR. The alignments obtained can be of three types (Examples refer to Figure \ref{fig:eg_amr}):

\begin{itemize}[noitemsep]
\item \textit{A single word aligned to a single concept}: E.g., word `read' aligned to concept `read-01'.
\item \textit{Span of words aligned to a graph fragment}: E.g., span `Stories from Nature' aligned to the graph fragment rooted at 'name'. This usually happens for named entities and multiword expressions such as those related to date and time.
\item \textit{A word aligned to \textsc{null} concept}: Most function words like `about', `a', `the', etc  are not aligned to any particular concept. These are considered to be aligned to the \textsc{null} concept. 
\end{itemize}

\textbf{Forced alignments}: The JAMR aligner does not align all concepts in a given AMR to a span in the sentence. We use a heuristic to forcibly align these leftover concepts and improve the quality of alignments. For every unaligned concept, we count the number of times an unaligned word occurs in the same sentence with the unaligned concept across all training examples. We then align every leftover concept in every sentence with the unaligned word in the sentence with which it has maximally coocurred.

\textbf{Span identification}: During training time, the aligner takes in a sentence and its AMR graph and splits each sentence into spans that can be aligned to the concepts in the AMR. However, during test time, we do not have access to the AMR graph. Hence, given a test sentence, we need to split the sentence into spans, on which we can predict concepts. We consider each word as a single span except for two cases. First, we detect possible multiword spans corresponding to named entities, using a named entity recognizer \cite{lafferty2001conditional}. Second, we use some basic regular expressions to identify time and date expressions in sentences.

\subsection{Experiments}

To train our model, we use \textsc{searn} as implemented in the Vowpal Wabbit machine learning library \cite{langford2007vowpal,daume2014efficient}. 

For each dataset, we run three kinds of experiments. They differ in how they get the concepts during test time. All of them use the approach described in Section \ref{subsec:learning} for predicting the relations. 

\begin{itemize}[noitemsep]
\item \textbf{Oracle Concept} - Use the true concept aligned with each span.
\item \textbf{1-Best Concept} - Use the concept with which the span was most aligned in the training data.
\item \textbf{Fully automatic} - Use the concepts predicted using the approach described in Section \ref{subsec:learning}.
\end{itemize}

\subsection{Connectivity}

Algorithm \ref{algo:learning} does not place explicit constraints on the structure of the AMR. Hence, the predicted output can have disconnected components. Since we want the predicted AMR to be connected, we connect the disconnected components (if any) using the following heuristic. For each component, we find its roots (i.e. concepts with no incoming relations). We then connect the components together by simply adding an edge from our predicted root $c_{root}$ to each of the component roots. To decide what edge to use between our predicted root $c_{root}$ and the root of a component, we get the $k$-best list (as described in section \ref{subsec:kbest}) between them and choose the most frequent edge from this list.

\subsection{Acyclicity}

\begin{table*}[!t]
\begin{center}
\begin{tabular}{|c|c|c|c|c|c|c|c|c|c|c|c|c|}
\hline
\textbf{Dataset}	& \multicolumn{9}{c}{\textbf{Our Results}} & \multicolumn{3}{|c|}{\textbf{JAMR Results}} \\
\hline
\multirow{2}{*}{}	& \multicolumn{3}{|c|}{Oracle Concepts} & \multicolumn{3}{|c|}{1-Best Concepts} & \multicolumn{3}{|c|}{Fully Automatic}  & \multicolumn{3}{|c|}{Fully Automatic} \\\cline{2-13} 
					& P & R & $F_1$ & P & R & $F_1$ & P & R & $F_1$ & P & R & $F_1$ \\		
\hline
bolt 			& 0.64 & 0.53 & 0.58 & 0.52 & 0.43 & \textbf{0.47} &  0.51 & 0.42 & \textbf{0.46} & 0.55 & 0.33 & \textbf{0.41} \\
proxy 		& 0.69 & 0.65 & 0.67 & 0.61 & 0.59 & \textbf{0.60} & 0.62 & 0.60  	& \textbf{0.61} & 0.68 & 0.53 & \textbf{0.59} \\ 
xinhua	 	& 0.68 & 0.60 & 0.64 &  0.55 & 0.49 & \textbf{0.52}  & 0.56 & 0.50 & \textbf{0.52}   &  0.59 & 0.40 & \textbf{0.48} \\ 
dfa		 	& 0.62 & 0.47 & 0.54 & 0.48 & 0.37 & \textbf{0.42} & 0.48 & 0.40 & \textbf{0.44}  & 0.52 & 0.15 & \textbf{0.23} \\ 
lp			& 0.70 & 0.58 & 0.63  & 0.57 & 0.45 & \textbf{0.50} & 0.54 & 0.49 & \textbf{0.52} & 0.53 & 0.41 & \textbf{0.46} \\ 
\hline
\end{tabular}
\end{center}
\caption{Full Results}
\label{tab:resultmain}
\end{table*}

\begin{table*}[!t]
\begin{center}
\begin{tabular}{|c|c|c|c|c|c|c|c|c|c|}
\hline
\textbf{Dataset}	& \multicolumn{6}{c}{\textbf{Our Results}} & \multicolumn{3}{|c|}{\textbf{JAMR Results}} \\
\hline
\multirow{2}{*}{}	& \multicolumn{3}{|c|}{1-Best} & \multicolumn{3}{|c|}{Fully Automatic}  & \multicolumn{3}{|c|}{Fully Automatic} \\\cline{2-10} 
					& P & R & $F_1$ & P & R & $F_1$  & P & R & $F_1$ \\
\hline
bolt 				& 0.74 & 0.72 & \textbf{0.73} & 0.74 & 0.72 & \textbf{0.73} &  0.73 & 0.55 & \textbf{0.63} \\
proxy 			& 0.79 & 0.77 & \textbf{0.78} & 0.78 & 0.78 & \textbf{0.78} &  0.78 & 0.68 & \textbf{0.73} \\ 
xinhua	 		& 0.74 & 0.77 & \textbf{0.74} & 0.74 & 0.77 & \textbf{0.75} & 0.69 & 0.57 & \textbf{0.63} \\ 
dfa		 		& 0.76 & 0.72 & \textbf{0.74} & 0.74 & 0.76 & \textbf{0.75}  &  0.85 & 0.33 & \textbf{0.48} \\ 
lp				& 0.77 & 0.79 & \textbf{0.78} & 0.77 & 0.80 & \textbf{0.79}  & 0.53 & 0.41 & \textbf{0.46} \\ 
\hline
\end{tabular}
\end{center}
\caption{Concept Prediction Results}
\label{tab:conceptresults}
\end{table*}

The post-processing step described in the previous section ensures that the predicted AMRs are rooted, connected, graphs. However, an AMR, by definition, is also acyclic. We do not model this constraint explicitly within our learning framework. Despite this, we observe that only a very small number of AMRs predicted using our fully automatic approach have cycles in them. Out of the total 1,444 AMRs predicted in all test sets, less than 5\% have cycles in them. Besides, almost all cycles that are predicted consist of only two nodes, i.e. both ${r_{ij}}$ and ${r_{ji}}$ have non-\textsc{no-edge} values for concepts $c_i$ and $c_j$. To get an acyclic graph, we can greedily select one of $r_{ij}$ or $r_{ji}$, without any loss in parser performance.

\subsection{Results}

Table \ref{tab:resultmain} shows the result of running our parser on all five datasets. By running our fully automatic approach, we get an absolute improvement of about \textbf{2\% to 6\%} on most datasets as compared to JAMR. Surprisingly, we observe a large improvement of \textbf{21\%} on the online discussion forum dataset (dfa). In all cases, our results indicate a more balanced output in terms of precision and recall as compared to JAMR, with consistently higher recall.

It should be noted that selecting the 1-best concept also gives better results than JAMR. This indicates that the 1-best baseline is strong, and possibly, not very easy to beat. To reinforce this, we evaluate our concept predictions separately. The results are shown in Table \ref{tab:conceptresults}. First, observe that going from the fully learned concept prediction to the 1-best concept shows only a small (or in some cases, no) drop in performance.  Second, note that we show a consistent absolute improvement of  \textbf{10\% to 12\%} over the concept prediction results of JAMR. As in the full prediction case, we observe a large performance increase (\textbf{27\%}) on the online discussion forum dataset.

\section{Related work}

Semantic representations and techniques for parsing them have a rich and varied history. AMR itself is based on propositional logic and neo-Davidsonian semantics \cite{davidson1967logical}. AMR is not intended to be an interlingua, but due to the various assumptions made while creating an AMR (dropping tense, function words, morphology, etc.), it does away with language-specific idiosyncrasies and interlingual representations \cite{dorr1992use} are thus, important predecessors to AMR. 

 Like the task of AMR parsing, there have been various attempts to parse sentences into a logical form, given raw sentences annotated with such forms \cite{DBLP:conf/aaai/KateWM05,DBLP:conf/naacl/WongM06}. The work by Zettlemoyer and Collins \cite{DBLP:conf/uai/ZettlemoyerC05} attempts to map natural language sentences to a lambda-calculus encoding of their semantics. They do so by treating the problem as a structured learning task, and use a log-linear model to learn a Probabilistic \emph{Combinatory Categorical Grammar} (CCG) \cite{steedman2011combinatory}, which is a grammar formalism based on lambda calculus. 

AMR aims to combine various semantic annotations to produce a unified annotation, but it mainly builds on top of PropBank \cite{DBLP:journals/coling/PalmerKG05}. PropBank has found extensive use in other semantic tasks such as shallow semantic parsing \cite{DBLP:conf/acl/GiugleaM06},

In our work we used \textsc{searn} to build an AMR parser. \textsc{searn} comes from a family of algorithms called "Learning to Search (L2S)" that solves structured prediction problems by decomposing the structured output in terms of an explicit search space and then learning a policy that can take actions in this search space in the optimal way. Incremental structured perceptron \cite{collins2004incremental,huang2012structured},
\textsc{DAgger} \cite{ross11dagger},
\textsc{Aggrevate} \cite{ross2014reinforcement}, etc. \cite{daume2005learning,xu2007learning,xu2007discriminative,ratliff2007boosting,syed2010reduction,doppa2012output} are other algorithms that also belong to this family. 

\section{Conclusion and Future work}

We have presented a novel technique for parsing english sentences into AMR using a learning to search approach. We model the concept and the relation learning in a unified framework using \textsc{searn} which allows us to optimize over the loss of the entire predicted output. We evaluate our parser on multiple datasets from varied domains and show that our parser performs better than the state-of-the-art across all the datasets. We also show that a simple technique of choosing the most frequent concept gives us a baseline that is better than the state-of-the-art for concept prediction. 

Currently we ensure various properties of AMR, such as connectedness and acyclicity using  heuristics. In the future, we plan to incorporate these as constraints in our learning technique. 

\bibliographystyle{acl}
\bibliography{AMR_Parsing_combined}
\end{document}